\documentclass[runningheads]{llncs}
\usepackage[T1]{fontenc}
\usepackage[utf8]{inputenc}
\usepackage{graphicx}
\usepackage{booktabs}
\usepackage[misc]{ifsym}
\usepackage{amssymb}
\usepackage{amsmath}
\usepackage{bm}
\usepackage{xcolor}
\usepackage{siunitx}
\usepackage{underscore}
\usepackage{esvect}
\usepackage{multirow}
\usepackage{natbib}
\setcitestyle{numbers}
\usepackage{hyperref}
\usepackage{tikz}
\usetikzlibrary{calc,positioning,arrows.meta}
\newcommand{\corr}{(\Letter)}

\begin{document}

\title{Competition-Aware CPC Forecasting with Near-Market Coverage}
\titlerunning{Competition-Aware CPC Forecasting with Near-Market Coverage}

\author{
Sebastian Frey\textsuperscript{*} \and
Edoardo Beccari\textsuperscript{*} \and
Maximilian Kranz\textsuperscript{*} \and
Nicolò Alberto Pellizzari\textsuperscript{*} \and
Ali Mete Karaman\textsuperscript{*} \and
Qiwei Han\corr\textsuperscript{*} \and
Maximilian Kaiser\textsuperscript{\dag}
}

\authorrunning{S. Frey et al.}

\institute{
\textsuperscript{*}Nova School of Business and Economics, Portugal\\
\email{66172@novasbe.pt, 67990@novasbe.pt, 67080@novasbe.pt, 63585@novasbe.pt, 67799@novasbe.pt, qiwei.han@novasbe.pt}\\[4pt]
\textsuperscript{\dag}University of Hamburg, Germany\\
\email{maximilian.kaiser@uni-hamburg.de}
}

\toctitle{Competition-Aware CPC Forecasting with Near-Market Coverage}
\tocauthor{Sebastian Frey, Edoardo Beccari, Maximilian Kranz, Nicolò Alberto Pellizzari, Ali Mete Karaman, Qiwei Han, Maximilian Kaiser}

\maketitle

\begin{abstract}
Cost-per-click (CPC) in paid search is an auction-generated outcome shaped by a competitive landscape that is only partially observable from any single advertiser's history. From 1.66 billion Google Ads log records for a concentrated car-rental market (2021--2023), we construct a weekly panel of 1,811 keyword series over 127 weeks (218,924 keyword--week observations) and build competition-aware proxies from keyword text, CPC trajectories, and geographic market structure. The design combines (i) semantic neighborhoods and a semantic keyword graph from pretrained transformer-based keyword representations, (ii) behavioral neighborhoods from Dynamic Time Warping (DTW) alignment of CPC trajectories, and (iii) geographic-intent covariates capturing localized demand and marketplace heterogeneity. We evaluate these signals both as exogenous covariates and as relational priors in spatiotemporal graph forecasters, benchmarking them against statistical, neural, and time-series foundation-model baselines. The results reveal a clear horizon crossover. At one week, graph-based models achieve the lowest error, reducing sMAPE by 15.1\% relative to the strongest classical/ML baseline; at the six- and twelve-week horizons, covariate-augmented foundation models dominate, reducing sMAPE by 22.5\% and 27.6\%, respectively. The gains concentrate in the high-CPC, high-volatility keywords where forecasting errors are most costly. A falsification battery supports the competition interpretation at the planning horizon: the semantic competition graph outperforms a confounder-matched non-competitive graph by 4.05 sMAPE points, and matched-neighbour and time-shuffled controls show the six-week gains are competition-specific rather than generic smoothing. Together, the findings establish a horizon-dependent competition-aware forecasting design for auction-driven advertising markets under partial observability.

\end{abstract}

\section{Introduction}
\label{sec:introduction}

Paid search advertising allocates sponsored placements through real-time auctions triggered by user queries, and advertisers typically pay under a cost-per-click (CPC) pricing scheme \cite{laffey2007paid,evans2009online,Jansen2013}. In sponsored search, CPC is not a fixed tariff but an auction-generated outcome shaped by bid-dependent ranking, platform-estimated quality signals, and query-specific auction conditions \cite{milgrom2010simplified,feldman_algorithmic_2008,nekipelov2017inference}. For advertisers, this creates a difficult forecasting problem: CPC determines how much demand a fixed budget can buy, yet the competitive state that generates CPC is only partially observable from any single advertiser's data.

This partial observability is central. Advertisers observe realized CPC, clicks, impressions, and spend, but they do not observe competitor bids, competitor quality signals, or the full auction state underlying price formation \cite{feldman_algorithmic_2008}. Our objective is therefore not to recover the latent auction state structurally, but to approximate competition through observable proxies. This distinction matters because purely autoregressive approaches are often informative over short horizons, yet become less reliable when regime shifts, localized demand changes, and competitor reallocation increasingly shape CPC dynamics.

To address this limitation, we construct three complementary proxy families from observable data. Semantic similarity captures substitutable intent across keywords; behavioral alignment captures shared exposure to latent demand and auction shocks; and geographic intent captures segmented market context and localized competitive intensity. Operationally, we instantiate these ideas through semantic neighborhoods and a fixed semantic keyword graph, Dynamic Time Warping (DTW)-based behavioral neighborhoods, and hierarchical geographic-intent covariates. These proxy families allow us to study two distinct representation routes: competition-aware covariates for general forecasters, and relational priors for spatiotemporal graph neural networks (STGNNs).

We evaluate this design on large-scale Google Ads logs from the car-rental sector, a setting in which demand is geographically localized, acquisition costs are operationally important, and competition is concentrated among a relatively small set of major advertisers. After preprocessing, the data yield a weekly panel of 1,811 keywords over 127 weeks. This does not imply full market observability. Rather, it creates a setting with sufficiently broad realized-outcome coverage to learn structured competition proxies from repeated interactions while remaining fundamentally partially observed.

Across a broad model universe spanning classical baselines, neural baselines, time-series foundation models (TSFMs), and STGNNs, we find clear horizon dependence in the value of competition-aware forecasting design. Graph-based models perform best at the 1-week horizon, where short-run spillovers are most relevant. At the 6-week and 12-week horizons, covariate-augmented TSFMs become clearly dominant, with coarse geographic structure acting as a particularly strong stabilizing prior. These gains are concentrated in the most operationally consequential part of the market: high-CPC, high-volatility keywords.

The paper makes four contributions. First, it reframes keyword-level CPC forecasting as a problem of partial competition observability, arguing that auction-generated prices can be forecast more effectively when latent competition is approximated through observable proxy structure. Second, it operationalizes this idea at industrial scale by constructing semantic, behavioral, and geographic competition proxies and evaluating them across both covariate-based and graph-based forecasting routes. Third, it shows that the value of competition-aware design is strongly horizon-dependent: semantic relational structure is most useful for short-run spillovers, while coarse geographic context is most useful as a stabilizing prior at medium and longer horizons. Fourth, it stress-tests the competition interpretation with a falsification battery: matched non-competitive, random, and time-shuffled neighbour controls, a confounder-matched non-competitive graph, and per-region segmentation. At the six-week planning horizon the semantic competition graph outperforms the matched non-competitive graph by 4.05 sMAPE points ($p<10^{-39}$), showing the gains there are competition-specific rather than generic smoothing or feature enrichment, while the same battery delimits precisely the horizons and markets where the effect holds.

\section{Related Work}
\label{sec:related_works}

\paragraph{CPC and paid search auctions.}
In Generalized Second Price (GSP) auctions, ad slots are assigned by ranking advertisers on an effective score that combines bids with platform-estimated quality factors \cite{keerthi2007constructing,caragiannis2014bounding}. Realized CPC is therefore an endogenous auction outcome shaped by rival bids, click-through-rate predictions, and latent demand shifts \cite{aggarwal2008sponsored,feldman2008algorithmic}. A substantial literature explains how CPC is generated in sponsored search and how auction design and ranking affect advertiser outcomes \cite{yang2022clickthrough}. However, this literature is more informative about price formation than about forecasting under partial competition observability, where the competitive state is not directly observed from one advertiser's history.

\paragraph{Keyword-level CPC forecasting.}
Classical baselines such as ARIMA and Prophet remain common reference points in business forecasting \cite{Taylor2018}, including advertising applications \cite{oldenburg2024interpretable}, but they often struggle with nonlinear and non-stationary dynamics. More flexible machine-learning and deep-learning approaches can improve performance, yet typically depend on task-specific feature construction and do not directly address the problem that the competitive structure driving CPC is only indirectly observed. The closest adjacent empirical work is \cite{oldenburg2024interpretable}, which studies keyword-level CPC forecasting with competitive-landscape covariates, but does not evaluate foundation-style forecasters, does not construct text-derived relational structure, and does not compare covariate-based and graph-based competition representations within a unified framework.

\paragraph{Time-series foundation models.}
Recent time-series foundation models (TSFMs), including TimeGPT \cite{garza_timegpt-1_2024}, Chronos \cite{ansari_chronos_2024}, and Moirai \cite{woo2024moirai}, have demonstrated strong zero-shot and few-shot performance across diverse forecasting tasks. Their application to advertising-cost forecasting remains limited, and in particular little is known about how these pretrained forecasters interact with competition-aware exogenous covariates in auction-generated settings. This makes CPC forecasting a relevant testbed for understanding whether large pretrained models can absorb proxy-based competition structure without an explicit relational topology.

\paragraph{Graph-based forecasting and latent interaction structure.}
Graph neural networks have become standard tools for spatiotemporal forecasting in traffic, energy, and sensor systems, where graph structure is typically given exogenously by physical topology or network connectivity. CPC forecasting differs in a crucial respect: there is no natural physical graph. Dependence arises from substitutable intent, shared exposure to demand shocks, and localized market overlap rather than from explicit network links \cite{wurfel2021online,yang2022clickthrough}. In this setting, the central problem is not only forecasting on a graph, but constructing a competition graph and competition covariates when the interaction structure is latent and only indirectly observable.

\paragraph{Research gap.}
Three gaps follow. First, sponsored-search research provides rich auction-theoretic intuition but limited guidance for forecasting under partial competition observability. Second, advertising-cost forecasting has not yet systematically incorporated modern TSFMs in a keyword-level CPC setting. Third, existing graph-forecasting work offers limited evidence on how text-derived, behavior-derived, and geography-derived proxy structure should be represented when competition is latent rather than directly observed. Our contribution is therefore not a new universal forecasting architecture, but a competition-aware forecasting design that makes latent market structure usable across heterogeneous model families.

\section{Business Context and Data Characteristics}
\label{sec:business_data}

Paid search advertising is one of the most important performance channels in digital marketing, allocating sponsored placements through real-time auctions triggered by user queries \cite{laffey2007paid,Jansen2013}. In car rental, CPC shocks have direct operational consequences because acquisition costs can compress margins quickly in periods of elevated competition. The sector is a particularly suitable empirical setting because it combines intense competition among a limited set of major rental groups and intermediaries, high online price transparency, and strong geographic concentration around airports and major cities.

We build on a large-scale dataset of keyword-level Google Ads performance logs for the car-rental industry (2021--2023), provided by a European business intelligence company. The raw dataset contains approximately 1.66 billion log-level observations. Each record contains keyword and matched-query text, contextual information such as device category and search type, and core performance metrics including impressions, clicks, and advertising cost.

To construct a stable forecasting universe, we restrict attention to car-rental intent, normalize keywords to canonical forms, parse landing-page URLs into registrable domains, and apply a domain-quality filter that removes unrelated or extremely low-coverage sites. We then retain keywords observed in at least 110 distinct weeks out of a 127-week window. The final weekly panel contains 1,811 keyword series and 218,924 keyword--week observations.

Weekly CPC is defined as
\[
\texttt{cpc\_week}_{k,t} =
\frac{\texttt{adcost\_sum}_{k,t}}{\texttt{adclicks\_sum}_{k,t}}
\quad \text{for } \texttt{adclicks\_sum}_{k,t} > 0.
\]
This weekly aggregation provides a practical compromise between auction-level noise reduction and managerial relevance, as weekly CPC aligns naturally with campaign planning and budget-steering cycles.

The retained domain universe remains concentrated around a relatively small set of continuously active major brands and intermediaries. This does not imply full market observability. Rather, it creates broad realized-outcome coverage and repeated competitive interactions within the focal vertical, making the data dense enough to learn structured competition proxies while preserving the central informational asymmetry of the problem.

The resulting panel exhibits strong right skewness and heavy-tailed behavior typical of paid-search markets. For CPC, the weekly mean is 2.86 while the maximum reaches 80.16, with a pronounced upper tail (p99 = 12.13) and skewness of 3.34. Traffic and spend metrics are substantially more extreme, reflecting a Pareto-like structure in which a small set of highly competitive queries accounts for disproportionate volume \cite{clauset2009powerlaw,montgomery2014applied}. This heterogeneity motivates evaluation beyond average forecast error alone.

\begin{figure}[t]
\centering
\includegraphics[width=\linewidth]{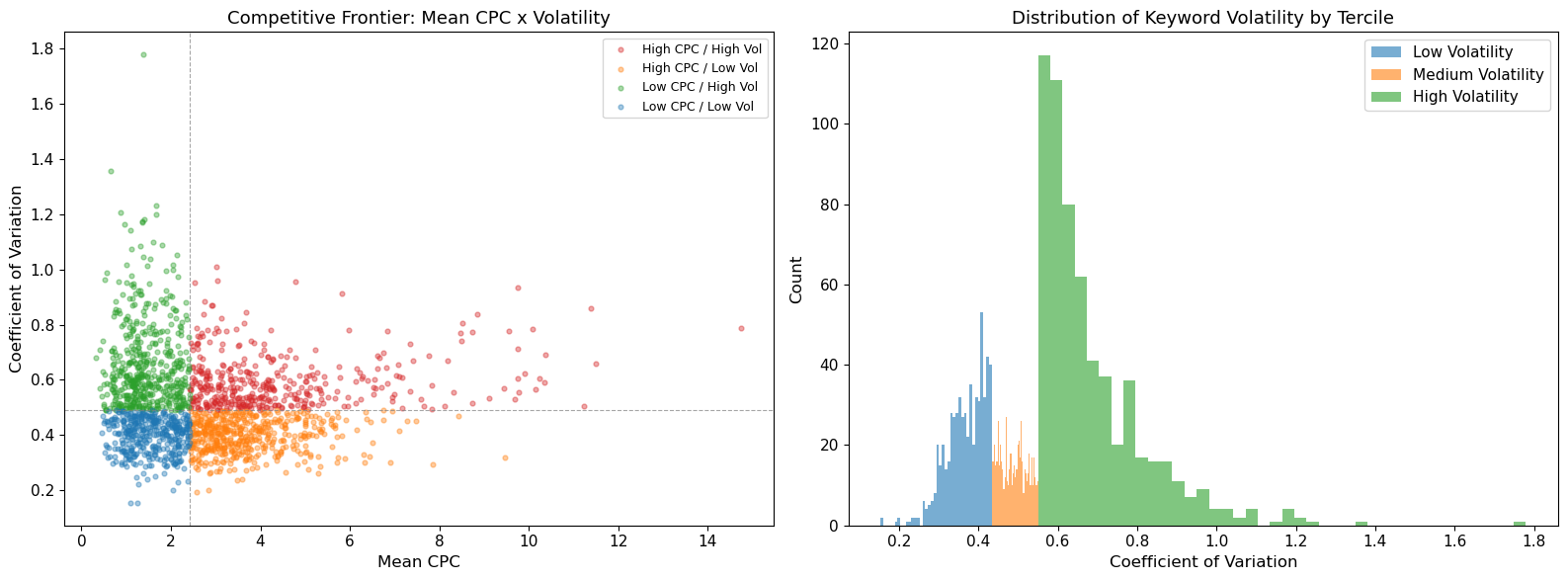}
\caption{Competitive frontier segmentation. The scatter plot maps keyword mean CPC against volatility (Coefficient of Variation), with median splits defining four distinct market regimes. The top-right quadrant identifies high-risk, high-value keywords.}
\label{fig:frontier_scatter}
\end{figure}

Figure~\ref{fig:frontier_scatter} summarizes this heterogeneity by mapping keyword mean CPC against the coefficient of variation. The resulting four quadrants define distinct market regimes, with the top-right quadrant representing the \emph{competitive frontier}: keywords that are both expensive and volatile. This region is strategically important because forecasting failures there carry disproportionate budget risk. It also provides a natural setting in which to test whether competition-aware proxy design delivers its largest gains where forecasting is hardest.

\section{Methodology: Competition-Aware Forecasting under Partial Observability}
\label{sec:methodology}

Keyword-level CPC forecasting in sponsored search is a prediction problem under partial observability. While advertisers observe realized CPC, impressions, clicks, and spend, they do not observe competitor bids, quality scores, budget constraints, or the full auction state that generates prices. Our objective is therefore not structural recovery of the auction mechanism, but scalable approximation of latent competition through observable proxy structure.

Our contribution lies in competition-proxy design and cross-family evaluation under partial observability, rather than in introducing a new universal forecasting backbone. As illustrated in Figure~\ref{fig:pipeline}, we operationalize latent competition through semantic, behavioral, and geographic signals, and study two implementation routes. In the first route, these proxies enter forecasting models as exogenous covariates. In the second, competition is represented relationally and supplied to spatiotemporal graph neural networks through a fixed semantic adjacency matrix. Across both routes, the target is weekly CPC at forecast horizon $h \in \{1,6,12\}$.

\begin{figure}
\begin{tikzpicture}[
    box/.style={
        draw, rounded corners, thick, align=center,
        inner sep=6pt, fill=white
    },
    topbox/.style={
        box, text width=10.4cm, fill=gray!10
    },
    proxybox/.style={
        box, text width=3.15cm
    },
    routebox/.style={
        box, text width=4.45cm
    },
    modelbox/.style={
        box, text width=4.65cm, fill=blue!5
    },
    outbox/.style={
        box, text width=10.0cm, fill=gray!10
    },
    arrow/.style={-{Latex[length=3mm]}, thick}
]
\node[topbox] (data) at (0,0) {\textbf{Observable Keyword Panel}\\CPC, Impressions, Clicks, Text};
\node[proxybox] (dtw) at (-4.6,-2.35) {\textbf{Behavioral Proxy}\\Shared CPC Dynamics};
\node[proxybox] (sem) at (0,-2.35) {\textbf{Semantic Proxy}\\Substitutable Intent};
\node[proxybox] (geo) at (4.6,-2.35) {\textbf{Geographic Proxy}\\Local Market Context};
\node[routebox] (route1) at (-2.2,-5.0) {\textbf{Route 1: Covariates}\\Lagged Proxy Features ($X$)};
\node[routebox] (route2) at ( 3.2,-5.0) {\textbf{Route 2: Relational Prior}\\Semantic Adjacency ($A$)};
\node[modelbox] (tsfm) at (-2.2,-7.25) {\textbf{Covariate-Augmented Models}\\Baselines and TSFMs};
\node[modelbox] (stgnn) at ( 3.2,-7.25) {\textbf{Graph Message Passing}\\STGNNs};
\node[outbox] (output) at (0,-9.65) {\textbf{Multi-Horizon Evaluation}\\$h \in \{1,6,12\}$ Weeks};
\draw[thick] (data.south) -- (0,-1.2);
\draw[arrow] (0,-1.2) -| (dtw.north);
\draw[arrow] (0,-1.2) -- (sem.north);
\draw[arrow] (0,-1.2) -| (geo.north);
\draw[thick] (dtw.south) -- (-4.6, -3.3);
\draw[thick] (sem.south) -- (0, -3.3);
\draw[thick] (geo.south) -- (4.6, -3.3);
\draw[thick] (-4.6, -3.3) -- (4.6, -3.3);
\draw[arrow] (-2.2, -3.3) -- (route1.north);
\draw[thick] (0, -3.3) -- (0, -3.9);
\draw[arrow] (0, -3.9) -| (route2.north);
\draw[arrow] (route1.south) -- (tsfm.north);
\draw[arrow] (route2.south) -- (stgnn.north);
\path (tsfm.south |- output.north) coordinate (outL);
\path (stgnn.south |- output.north) coordinate (outR);
\draw[arrow] (tsfm.south) -- (outL);
\draw[arrow] (stgnn.south) -- (outR);
\end{tikzpicture}
\caption{Competition-aware forecasting design under partial observability. Latent competition is approximated through three proxy families and represented either as exogenous covariates or as a fixed relational prior under a common multi-horizon evaluation protocol.}
\label{fig:pipeline}
\end{figure}

\subsection{Competition Proxy Extraction}
\label{subsec:proxy_extraction}

We derive three complementary proxy families from observable data. Each proxy is designed to approximate a different aspect of latent competition that is not directly observable from any single advertiser's history.

\paragraph{Semantic proxy: substitutable intent.}
Keywords with similar user intent often compete for overlapping advertising inventory even when they differ lexically. To capture this intent-level proximity, we encode each keyword $k_i$ using pretrained transformer-based text representations from \texttt{all-MiniLM-L6-v2} \cite{reimers2019sentencebert}. The language model is used only as a representational device: it supplies fixed semantic embeddings rather than generative outputs. This yields an embedding $e_i \in \mathbb{R}^{384}$. Pairwise cosine similarity is then used to identify semantically related keywords. These semantic neighborhoods serve two purposes: they generate neighborhood-based competition covariates and they define the topology of a fixed semantic keyword graph.

\paragraph{Behavioral proxy: shared exposure to latent shocks.}
Textual similarity does not exhaust competitive relatedness. Keywords with weak lexical overlap may still exhibit similar CPC dynamics because they are exposed to common demand shocks, campaign timing, or auction reallocation. We therefore construct behavioral neighborhoods using Dynamic Time Warping (DTW), which measures similarity between CPC trajectories while allowing local temporal misalignment \cite{Herrmann2023}. To avoid pathological alignments, we impose a Sakoe--Chiba band constraint. In the covariate route, these neighbors are summarized into leakage-free behavioral competition features constructed from historical CPC trajectories only.

\paragraph{Geographic proxy: segmented market context.}
In the car-rental setting, auction pressure is strongly localized. Queries tied to airports, cities, or countries reflect distinct demand pools and competitive conditions. We extract geographic intent from keyword text, curated gazetteers, and hierarchical geographic mapping. Each keyword is assigned structured location indicators at the continent, country, and city levels. Rather than assuming that finer geography is always better, we retain multiple hierarchical resolutions and evaluate them empirically. These geography-derived variables act as spatial competition proxies by encoding localized demand conditions and marketplace heterogeneity.

\subsection{Proxy Representation: Covariates and Relational Structure}
\label{subsec:proxy_representation}

The extracted proxies enter the forecasting problem through two different representations. The first route treats competition information as exogenous conditioning variables. These include leakage-free summaries of neighboring CPC histories derived from semantic and DTW neighborhoods, together with structured geographic-intent indicators and core operational controls. This route allows forecasting models to condition on observable approximations to latent competition without imposing an explicit graph topology.

The second route represents competition relationally. Using the keyword embeddings $e_i$, we construct a fixed semantic graph $A^{sem} \in \mathbb{R}^{N \times N}$ over the keyword universe. For each node, directed edges are assigned to its $k$ nearest semantic neighbors under cosine similarity, with $k=10$ selected as a stable and competitive graph sparsity level in preliminary experiments. The adjacency matrix is row-normalized before model training.

We restrict the relational route to a fixed semantic graph to isolate the effect of stable text-grounded competition structure and to maintain comparability across STGNN architectures. Behavioral proximity is incorporated instead as leakage-free covariates because its dynamic, trajectory-derived nature makes it less suitable as a fixed \emph{ex ante} adjacency prior.

\subsection{Forecasting Architectures}
\label{subsec:forecasting_architectures}

We evaluate three model classes: classical and neural non-graph baselines, covariate-augmented time-series foundation models, and spatiotemporal graph neural networks.

\paragraph{Classical and neural baselines.}
We benchmark a broad suite of traditional forecasting models, including statistical approaches (SARIMAX), tree-based machine learning models (XGBoost, Random Forest, LightGBM), deep neural networks (MLP, LSTM, GRU), and a tabular foundation model (TabPFN). These baselines establish the reference performance attainable from autoregressive and weakly contextual histories.

\paragraph{Covariate-augmented foundation models.}
We benchmark state-of-the-art TSFMs, including Chronos-2 \cite{amazon2024chronos2}, TimeGPT \cite{garza_timegpt-1_2024}, and Moirai \cite{woo2024moirai}, under exogenous conditioning. In this setting, competition proxies are supplied as covariates alongside the historical CPC series and core operational inputs. This route tests whether large pretrained forecasters can absorb competition-aware structure without an explicit relational topology.

\paragraph{Spatiotemporal graph neural networks.}
To model cross-keyword dependence explicitly, we evaluate STGNNs that consume the fixed semantic graph $A^{sem}$. The input panel is represented as a spatiotemporal tensor
\[
X \in \mathbb{R}^{N \times T \times F},
\]
where $N$ is the number of keywords, $T$ the input history length, and $F$ the number of features. The model outputs multi-horizon forecasts $\hat{Y}_{t+h} \in \mathbb{R}^{N}$.

We focus on three representative STGNN architectures. DCRNN models competition as diffusion over the predefined graph within a recurrent forecasting framework \cite{li_diffusion_2018}. GConvLSTM integrates graph convolutions directly into LSTM gating, providing a stable convolutional-recurrent formulation. GraphWaveNet combines fixed graph structure with an adaptive component, allowing limited refinement of the exogenous relational prior \cite{wu_graph_2019}. All STGNNs are trained globally on the CPC panel using a leakage-free chronological split and optimized with Mean Absolute Error (MAE), which is more robust than squared-error objectives under the heavy-tailed distribution of CPC.

\section{Evaluation Procedures and Results}
\label{sec:eval_results}

\subsection{Experimental Protocol}
All models are evaluated on weekly CPC forecasting for 1,811 keyword series under a strict chronological split, with the final 20\% of observations reserved for out-of-sample testing to prevent temporal leakage. We study three forecasting horizons, $h \in \{1,6,12\}$ weeks, corresponding respectively to short-term bid steering, medium-term tactical planning, and longer-range budget allocation.

We report Symmetric Mean Absolute Percentage Error (sMAPE) as the primary business metric and Root Mean Squared Error (RMSE) as a magnitude-sensitive secondary metric. Because CPC is heavily right-skewed and keyword heterogeneity is substantial, all results are aggregated across the keyword panel and reported as mean $\pm$ standard deviation. Beyond point estimates, we assess whether per-horizon improvements are statistically reliable across the keyword panel using the paired Wilcoxon signed-rank test, a bootstrap 95\% confidence interval on the mean sMAPE difference, and the per-keyword win rate. To support reproducibility despite proprietary data restrictions, we release the preprocessing, feature-construction, graph-construction, modelling, and evaluation code together with graph-construction parameters, random seeds, hyperparameter grids, and chronological split definitions.\footnote{\url{https://github.com/Sebastian-Frey/Competition-Aware-GNNs-for-TimeSeriesForecasting}}

\subsection{Cross-Horizon Family-Level Comparison}
\label{subsec:cross_horizon}

Table~\ref{tab:cross_horizon_summary} provides a high-level performance summary using the strongest representative from each model family. The dominant architecture changes systematically across horizons. At the 1-week horizon, STGNNs perform best, indicating that relational structure is especially useful for short-run competitive spillovers. At the 6-week and 12-week horizons, covariate-augmented TSFMs dominate, suggesting that stable contextual priors, especially geographic ones, become more valuable as forecast distance increases. Across all three horizons, competition-aware families outperform classical and non-graph ML baselines. Table~\ref{tab:full_model_universe} reports the full model universe across all three horizons, providing the per-architecture basis for the family-level representatives in Table~\ref{tab:cross_horizon_summary} and confirming the horizon-dependent crossover: STGNNs lead at one week, while covariate-augmented TSFMs lead at six and twelve weeks. The graph-over-classical advantage is also statistically robust at the per-keyword level when comparing the best STGNN against the strongest classical/ML baseline at each horizon (Table~\ref{tab:added_value}).

\begin{table}[t]
\caption{Cross-horizon summary using the strongest representative from each model family. Reported metric is overall sMAPE (\%). Lower is better. The strongest representative may differ across horizons.}
\label{tab:cross_horizon_summary}
\centering
\begin{tabular}{@{}lccc@{}}
\toprule
\textbf{Model Family Representative} & \textbf{1-Week} & \textbf{6-Week} & \textbf{12-Week} \\
\midrule
Best Classical / ML Baseline & 30.42 & 35.04 & 40.23 \\
Best Covariate-Augmented TSFM & 26.66 & 27.14 & 29.14 \\
Best Spatiotemporal GNN & 25.82 & 30.42 & 37.46 \\
\bottomrule
\end{tabular}
\end{table}

\begin{table}[t]
\caption{Comprehensive multi-horizon evaluation across all baseline, foundation, and graph-based models. Metrics reflect the best-performing configuration for each model at each horizon. Reported value is mean sMAPE (\%).}
\label{tab:full_model_universe}
\centering
\begin{tabular}{@{}llccc@{}}
\toprule
\textbf{Model Family} & \textbf{Architecture} & \textbf{1-Week} & \textbf{6-Week} & \textbf{12-Week} \\
\midrule
\multirow{8}{*}{Classical \& ML Baselines}
& SARIMAX      & 43.74 & 43.93 & 44.30 \\
& XGBoost      & 30.71 & 36.64 & 40.91 \\
& LightGBM     & 30.68 & 36.62 & 40.89 \\
& RF           & 31.53 & 38.25 & 43.02 \\
& MLP          & 30.42 & 38.10 & 40.23 \\
& LSTM         & 30.70 & 36.62 & 40.39 \\
& GRU          & 31.00 & 36.45 & 40.59 \\
& TabPFN       & 31.45 & 35.04 & 40.79 \\
\midrule
\multirow{3}{*}{Covariate-Augmented TSFMs}
& Moirai       & 34.22 & 30.14 & 34.28 \\
& TimeGPT      & 26.66 & 29.29 & 31.77 \\
& Chronos-2    & 27.94 & 27.14 & 29.14 \\
\midrule
\multirow{4}{*}{Spatiotemporal GNNs}
& GraphWaveNet & 25.82 & 30.57 & 37.81 \\
& GConvLSTM    & 26.06 & 30.69 & 38.33 \\
& DCRNN        & 26.06 & 30.42 & 37.46 \\
& AGCRN        & 26.79 & 30.76 & 37.77 \\
\bottomrule
\end{tabular}
\end{table}

\begin{table}[t]
\caption{Added value of the best graph model over the best classical/ML baseline at each horizon. The model choices match the per-horizon representatives in Table~\ref{tab:cross_horizon_summary}, but the GNN and classical sMAPE columns are re-aggregated over the $1{,}809$ keywords common to both compared models, so $\Delta$ equals their difference. $\Delta$ is the paired per-keyword mean difference in sMAPE (negative favours the graph model), with a bootstrap 95\% confidence interval; all paired differences are significant under a Wilcoxon signed-rank test ($p<10^{-20}$).}
\label{tab:added_value}
\centering
\begin{tabular}{@{}llccccc@{}}
\toprule
\textbf{Horizon} & \textbf{Model pair} & \textbf{GNN} & \textbf{Base} & \textbf{$\Delta$} & \textbf{95\% CI} & \textbf{Win} \\
\midrule
1-week  & GraphWaveNet vs.\ MLP & 25.77 & 30.42 & $-4.65$ & $[-5.13,\,-4.19]$ & 82\% \\
6-week  & DCRNN vs.\ TabPFN     & 30.37 & 35.05 & $-4.68$ & $[-5.47,\,-3.91]$ & 65\% \\
12-week & DCRNN vs.\ MLP        & 37.38 & 40.22 & $-2.84$ & $[-4.34,\,-1.27]$ & 62\% \\
\bottomrule
\end{tabular}
\end{table}

\subsection{Main Comparison at the 6-Week Planning Horizon}
\label{subsec:main_comparison}

We focus the main detailed comparison on the 6-week horizon because it is both managerially central and methodologically revealing. Relative to the 1-week horizon, it is less dominated by immediate short-run spillovers. Relative to the 12-week horizon, it is less affected by accumulated long-range drift. It therefore provides the most decision-relevant tactical planning horizon and the clearest separation among model families.

Table~\ref{tab:grand_comparison_6w} reports the strongest configurations within each architecture family at this horizon.

\begin{table}[t]
\caption{Main comparison of forecasting architectures at the 6-week horizon. Metrics are Mean $\pm$ Standard Deviation across 1,811 keywords. Bold values indicate the best result within each model family and globally.}
\label{tab:grand_comparison_6w}
\centering
\resizebox{\textwidth}{!}{%
\begin{tabular}{@{}l l l c c@{}}
\toprule
\textbf{Model Family} & \textbf{Architecture} & \textbf{Best Competition-Aware Setup} & \textbf{sMAPE (\%)} & \textbf{RMSE} \\
\midrule
\multirow{3}{*}{\shortstack[l]{Statistical \& ML\\Baselines}}
& SARIMAX & Univariate lags & 43.93 $\pm$ 23.55 & 1.660 $\pm$ 1.759 \\
& XGBoost & Core operational features & 36.64 $\pm$ 17.51 & 1.301 $\pm$ 1.119 \\
& \textbf{TabPFN (1-shot)} & \textbf{Core operational features} & \textbf{35.04 $\pm$ 17.77} & \textbf{1.250 $\pm$ 1.133} \\
\midrule
\multirow{3}{*}{\shortstack[l]{Covariate-Augmented\\TSFMs}}
& Moirai & Leakage-free lags + calendar stabilization & 30.14 $\pm$ 18.24 & 1.000 $\pm$ 0.970 \\
& TimeGPT & Calendar conditioning + growth clamp & 29.29 $\pm$ 17.07 & 1.002 $\pm$ 1.008 \\
& \textbf{Chronos-2} & \textbf{Geographic intent covariates} & \textbf{27.14 $\pm$ 15.04} & \textbf{0.841 $\pm$ 0.846} \\
\midrule
\multirow{3}{*}{\shortstack[l]{Spatiotemporal GNNs\\Relational Priors}}
& GraphWaveNet & Semantic graph + search mix & 30.57 $\pm$ 20.57 & 1.005 $\pm$ 0.941 \\
& GConvLSTM & Semantic graph + continent geography & 30.69 $\pm$ 20.42 & 1.001 $\pm$ 0.955 \\
& \textbf{DCRNN} & \textbf{Semantic graph + geo + semantic CPC} & \textbf{30.42 $\pm$ 20.42} & \textbf{1.000 $\pm$ 0.926} \\
\bottomrule
\end{tabular}%
}
\end{table}

Three patterns emerge. First, isolated or weakly contextual keyword histories reach a clear performance ceiling: the strongest baseline, TabPFN, attains 35.04\% sMAPE, while SARIMAX remains substantially worse. Second, competition-aware TSFMs lower the global error floor materially, with Chronos-2 using geographic-intent covariates reaching 27.14\% sMAPE, the strongest overall result. This indicates that coarse geographic localization acts as a powerful stabilizing prior for pretrained forecasters. Third, explicit relational modeling remains competitive and clearly improves on the non-graph baseline family: DCRNN, leveraging the semantic keyword graph alongside semantic-neighbor CPC signals, reaches 30.42\% sMAPE and provides a structurally distinct route to competition-aware forecasting.

\paragraph{Geographic resolution.}
Because geographic intent is one of the strongest competition-aware priors, we examine whether finer geographic granularity improves forecasting. The answer is consistently no. As shown in Table~\ref{tab:geo_resolution}, continent-level encoding outperforms country- and city-level encodings across all three horizons when averaged across graph backbones, by 0.61 sMAPE points over country and 0.92 over city at the 6-week horizon. The reason is statistical stability rather than semantic richness: coarse geography captures durable regional differences in demand intensity and seasonality without fragmenting the training signal.

\begin{table}[t]
\caption{Effect of geographic resolution averaged across graph backbones (mean sMAPE; lower is better). Coarse representations consistently yield the lowest error.}
\label{tab:geo_resolution}
\centering
\begin{tabular}{@{}c c c c@{}}
\toprule
Horizon & Continent (7 dummies) & Country (63 dummies) & City (268 dummies) \\
\midrule
1w  & 26.36 & 26.72 & 27.16 \\
6w  & 30.90 & 31.51 & 31.82 \\
12w & 37.93 & 38.70 & 39.04 \\
\bottomrule
\end{tabular}
\end{table}

\subsection{Competitive Frontier Analysis}
\label{subsec:frontier_analysis}

To assess where competition-aware proxy design matters most, we conduct a structured ablation across the GNN family and segment the keyword panel by CPC intensity and volatility. This segmentation yields well-populated partitions, with the high-CPC / high-volatility \emph{competitive frontier} quadrant containing 402 strategically critical keywords.

The ablation results show that proxy effectiveness depends strongly on both horizon and market regime. At the 6-week horizon, the strongest specification combines geography with semantic-neighbor CPC information (\texttt{Core + Geo + Sem CPC}), improving overall sMAPE from 31.61\% to 30.71\%. More importantly, this combination lowers error in the highly unstable, high-volatility segment by 1.3 percentage points relative to core-only inputs. At 12 weeks, the simpler \texttt{Core + Continent} setup becomes the strongest prior, reducing overall error from 38.32\% to 37.93\%, highlighting that coarse geography serves as a robust anchor against long-term drift.

The results also show that proxy design must be selective. In our setting, indiscriminate feature stacking is consistently counterproductive in noisy auction environments. At the 6-week horizon, naive aggregation of all available proxies produces the weakest performance (34.0\% sMAPE, 3.3 points worse than the strongest selective setup); at 12 weeks, this gap widens further. Taken together, these findings provide strong evidence that competition-aware proxy design is especially valuable in the hardest-to-forecast regions of the market, and that structured augmentation is preferable to exhaustive accumulation of all available signals.

\subsection{Robustness: Are the Gains Competition-Specific?}
\label{subsec:falsification}

A central concern is whether the gains from competition proxies reflect genuine competitive structure or merely generic contextual smoothing and feature enrichment. We address this with a falsification battery applied directly to the graph model: holding the architecture, training protocol, and evaluation fixed, we replace each competition signal with a null control and measure the change in per-keyword sMAPE. We report mean paired differences (real minus control; negative favours the real competition signal) with Wilcoxon signed-rank tests and bootstrap 95\% confidence intervals over the $1{,}811$ keywords.\footnote{These robustness experiments are interpreted as within-diagnostic contrasts: each real proxy and its null control are evaluated under the same chronological split, implementation, architecture, and training protocol. The table therefore reports relative paired effects rather than replacing the main benchmark estimates in Table~\ref{tab:grand_comparison_6w}.}

\paragraph{Neighbour and graph controls.} At the six-week planning horizon, the falsification evidence is strongest for the matched and time-shuffled controls, while the random-neighbour comparison is smaller and should be interpreted as marginal evidence (Table~\ref{tab:falsification}). The semantic-neighbour competition covariate outperforms a confounder-matched but non-competitive control by $1.99$ sMAPE points ($p<10^{-16}$), a random-neighbour control by $0.25$ ($p<0.05$), and a time-shuffled version of the same neighbours by $1.69$ ($p<10^{-21}$); removing the covariate entirely costs $0.82$ points ($p<10^{-12}$). The relational route is more nuanced. The matched non-competitive graph is a \emph{hard negative}: it links each keyword to neighbours that match on confounders normally associated with relatedness, such as mean CPC, volatility, and volume, but are semantically dissimilar. Its edges therefore look informative while connecting series that do not co-move, so message passing propagates a \emph{structured, misleading} prior; this degrades accuracy by $4.05$ points ($p<10^{-39}$). A random graph, by contrast, imposes no systematic relationship: its edges carry no consistent signal and average out as noise, leaving it on par with the semantic graph (within $0.6$ sMAPE). The contrast indicates that the damage comes from structured but wrong edges rather than from the absence of true competitive edges: an actively misleading prior hurts, a neutral one does not. At this horizon, then, the competition-specific signal resides primarily in the covariate route. This does not contradict the short-horizon result in Table~\ref{tab:cross_horizon_summary}; it clarifies the horizon dependence of the graph route: relational priors add value at one week, while medium-horizon gains are driven primarily by structured covariates and contextual stabilization. The advantage over the time-shuffled control further indicates that not only \emph{which} keywords are related but also the \emph{temporal alignment} of their histories carries competitive information.

\paragraph{Horizon dependence and scope.} The falsification evidence is horizon-specific. At one week, where short-run autoregressive dynamics are strong, the null controls are statistically difficult to separate from the real signal even though graph forecasters perform best overall. At twelve weeks, the relationship reverses within the graph diagnostic: competition covariates and the semantic graph begin to over-condition on noisy long-range competitive signals. Removing the semantic covariate, for instance, improves sMAPE by $2.81$ points ($p<10^{-26}$). This pattern explains why relational priors are most useful for short-run spillovers, while medium- and long-horizon performance is better stabilized by structured covariates, especially coarse geographic priors.

\paragraph{Geographic heterogeneity.}
Segmenting the geographic effect by continent shows it is not uniform: continent-level geography significantly helps European keywords ($-1.20$, $p<0.01$) and the unlocalised ``Global'' segment ($-0.71$, $p<0.05$), but \emph{hurts} North American keywords ($+0.48$, $p<0.01$). This heterogeneity is consistent with a market-specific effect rather than a purely generic smoother, and it delimits where geographic conditioning is beneficial.

\begin{table}[t]
\caption{Falsification of competition signals at the six-week horizon. Each entry is the mean paired difference in per-keyword sMAPE (\%) between the real competition signal and a null control (negative favours real competition), with a bootstrap 95\% confidence interval and Wilcoxon $p$-value ($n=1{,}811$).}
\label{tab:falsification}
\centering
\begin{tabular}{@{}lccc@{}}
\toprule
\textbf{Control} & \textbf{$\Delta$ sMAPE} & \textbf{95\% CI} & \textbf{$p$} \\
\midrule
Real vs.\ matched non-competitive neighbours & $-1.99$ & $[-2.36,\,-1.62]$ & $<10^{-16}$ \\
Real vs.\ random neighbours                  & $-0.25$ & $[-0.52,\,+0.02]$ & $<0.05$ \\
Real vs.\ time-shuffled neighbours           & $-1.69$ & $[-2.00,\,-1.39]$ & $<10^{-21}$ \\
Competition covariate on vs.\ off            & $-0.82$ & $[-1.08,\,-0.56]$ & $<10^{-12}$ \\
Semantic vs.\ matched non-competitive graph  & $-4.05$ & $[-4.57,\,-3.50]$ & $<10^{-39}$ \\
\bottomrule
\end{tabular}
\end{table}

\section{Conclusion}
\label{sec:conclusion}

This paper framed CPC forecasting in paid search as a problem of partial competition observability. Advertisers do not observe competitor bids, quality signals, or the full auction state, yet their realized CPC histories carry exploitable traces of competitive structure. Our proxies extract that structure at scale and forecast CPC substantially better than competition-blind baselines: competition-aware foundation-model specifications cut sMAPE by 22.5\% and 27.6\% at the six- and twelve-week horizons relative to the strongest classical/ML baseline, and our falsification battery confirms the six-week signal is competition-specific rather than generic smoothing.

The main lesson is not that one forecasting architecture uniformly dominates, but that under partial competition observability, different proxy families and representation routes become useful at different forecast horizons. Graph-based models deliver the strongest short-horizon performance, indicating that semantic relational structure is especially useful for capturing short-run competitive spillovers. At medium and longer horizons, covariate-augmented TSFMs become clearly dominant, with coarse geographic context acting as a particularly effective stabilizing prior. These gains are most meaningful in the part of the market where forecasting failures are most costly. Improvements concentrate in high-CPC, high-volatility keywords, where auction pressure is strongest and budget risk is highest. In this sense, competition-aware forecasting does not merely improve average accuracy; it improves robustness in the most operationally consequential segment of the market.

Several limitations remain. The analysis is based on a single vertical with a relatively concentrated competitive landscape, so generalization to broader or less concentrated paid-search categories remains an open question. In addition, the semantic graph is fixed over time and therefore cannot capture evolving keyword relationships, market entry and exit, or shifting competitive regimes. Future work should examine dynamic graph construction, richer competitor-side signals where available, and validation across additional industries and auction settings. Our falsification analysis (Section~\ref{subsec:falsification}) further shows that the competition signal is concentrated at the medium horizon and heterogeneous across markets, attenuating, and at the twelve-week horizon reversing, rather than improving uniformly; these are scope conditions of competition-aware design rather than universal guarantees. Finally, all evidence here is offline: because a deployed forecaster can itself influence bidding and hence future CPC---for example, budget reallocation by losing bidders can alter pacing and realized CPC---an online A/B evaluation is an important next step that an observational study cannot provide.

Overall, the findings show that auction-driven advertising markets can be forecast more effectively when competitive structure is modeled explicitly rather than treated as unobserved noise. The contribution is a horizon-dependent competition-aware forecasting design: relational priors are most valuable for short-run spillovers, while structured covariates, especially coarse geographic priors, stabilize medium- and long-horizon forecasts, turning partially observed competition into actionable forecasting signal without requiring bid-level access to the hidden auction state.

\section{Acknowledgements}
This work was supported by Funda\c{c}\~ao para a Ci\^encia e a Tecnologia (UID/00124/2025 and UID/PRR/124/2025; DOI: \url{https://doi.org/10.54499/UID/00124/2025}) and by LISBOA2030 (DataLab2030, LISBOA2030-FEDER-01314200).

\bibliographystyle{splncs04}
\bibliography{mybibliography}

\end{document}